\def\paperlanguage{}
\pdfoutput=1 

\documentclass[letterpaper, 10 pt, conference]{ieeeconf}  

\usepackage{bm}
\usepackage{cite}
\include{preamble}

\newcommand{\ctext}[1]{\raise0.2ex\hbox{\textcircled{\scriptsize{#1}}}}

\IEEEoverridecommandlockouts                              

\title{\LARGE \textbf
  {
    \switchlanguage%
    {%
      Diffusion-Based Body Schema Learning Enabling Abnormal-State Adaptation in Musculoskeletal Robots
    }%
    {%
      筋骨格ロボットにおける異常状態適応を可能にする\\拡散モデル型身体図式学習
    }%
  }
}

\author{Kento Kawaharazuka$^{1}$ and Shuhei Ikemoto$^{2}$
  \thanks{$^{1}$ The author is with the Department of Mechano-Informatics, Graduate School of Information Science and Technology, The University of Tokyo, 7-3-1 Hongo, Bunkyo-ku, Tokyo, 113-8656, Japan.
    {\texttt\small kawaharazuka@jsk.imi.i.u-tokyo.ac.jp}
  }
  \thanks{$^{2}$ The author is with the Graduate School of Life Science and Systems Engineering, Kyushu Institute of Technology, 7230, Hibikino 2-4, Wakamatsu, Kitakyushu, Fukuoka, 808-0135, Japan.
    {\texttt\small ikemoto@brain.kyutech.ac.jp}
  }
}

\begin{document}

\maketitle
\thispagestyle{empty}
\pagestyle{empty}

\begin{abstract}
  \switchlanguage%
  {%
    Musculoskeletal robots require an internal body schema that remains consistent under a wide range of physical state changes, including abnormalities such as muscle rupture and actuator jamming.
    Conventional approaches based on autoencoders or variational autoencoders learn average behaviors by projecting sensor and actuator signals into a low-dimensional latent space; however, exploration within the latent space alone has limited capability to handle out-of-distribution or abnormal states that are not included in the training data.
    To address this limitation, this study proposes a diffusion-based framework for body schema learning in musculoskeletal robots.
    Unlike generative models that operate through low-dimensional latent spaces, diffusion models can directly and iteratively estimate physically consistent sensor and actuator values in the high-dimensional space through a denoising process, even under partial observations and constraints, without requiring retraining.
    By formulating body schema adaptation as a gradient-guided denoising process, the proposed method enables adaptive estimation of appropriate muscle lengths and muscle tensions even under abnormal conditions such as muscle rupture and actuator jamming.
    The validity of the proposed framework is verified through simulation experiments using a musculoskeletal robot model.
    }%
  {%
    筋骨格ロボットには, 筋の断裂やアクチュエータの固着といった異常を含む, 広範な物理状態変化の下でも, 一貫性を保つ内部身体図式が求められる.
    オートエンコーダや変分オートエンコーダに基づく従来手法は, センサおよびアクチュエータ信号を低次元な潜在空間へ射影することで平均的な挙動を学習するが, 潜在空間の探索だけでは, 学習データに含まれない分布外状態や異常状態への対応能力には限界がある.
    そこで本研究では, 筋骨格ロボットにおける身体図式学習のための拡散モデルに基づく枠組みを提案する.
    低次な潜在空間を通した生成モデルとは異なり, 拡散モデルはノイズ除去過程によって, 再学習なしに, 部分観測や制約条件の下でも物理的整合性の取れたセンサ・アクチュエータ値を高次元空間で直接反復的に推定することができる.
    身体図式の適応を勾配に基づくガイダンスとノイズ除去の過程として定式化することで, 本手法は筋の断裂やアクチュエータの固着といった異常条件下においても, 適切な筋長や筋張力を適応的に推定することを可能にする.
  }%
\end{abstract}


\section{INTRODUCTION}\label{sec:introduction}
\switchlanguage%
{%
  To date, a wide variety of musculoskeletal robots that imitate diverse human functions and control mechanisms have been proposed \cite{gravato2010ecce1, jantsch2013anthrob, asano2016kengoro, kawaharazuka2019musashi}.
  Because these robots are based on human anatomical models, they exhibit various characteristics such as muscle redundancy, nonlinear elasticity, and anisotropy \cite{kawaharazuka2026muscle}.
  This enables variable stiffness control at the hardware level \cite{jacobsen1990control, koganezawa1999stiffness}, as well as the direct implementation and experimental validation of human reflex structures \cite{endo1994reflex, marques2013reflex, liu2018stretch}.
  Among these features, high adaptability arising from muscle redundancy is one of the greatest advantages of musculoskeletal robots.
  They are expected to continue operating in a human-like manner by adapting to abnormal conditions such as muscle rupture and actuator jamming \cite{kawaharazuka2022redundancy}.
}%
{%
  これまで, 人間のあらゆる機能や制御を模倣した, 様々な筋骨格ロボットが提案されてきた\cite{gravato2010ecce1, jantsch2013anthrob, asano2016kengoro, kawaharazuka2019musashi}.
  これらは人体模型ゆえに筋の冗長性や非線形弾性, 異方性などの様々な特徴を持ち\cite{kawaharazuka2026muscle}, ハードウェアレベルでの可変剛性制御\cite{jacobsen1990control, koganezawa1999stiffness}や, 人間の反射構造の直接的な実装と検証などが可能である\cite{endo1994reflex, marques2013reflex, liu2018stretch}.
  その中でも特に, 筋の冗長性を活かした高い適応能力は, 筋骨格ロボットの持つ最大の利点の一つである.
  筋破断やアクチュエータの固着といった異常状態に適応して, 人間同様に継続的に動作することが期待されている\cite{kawaharazuka2022redundancy}.
}%

\begin{figure}[t]
  \centering
  \includegraphics[width=0.95\columnwidth]{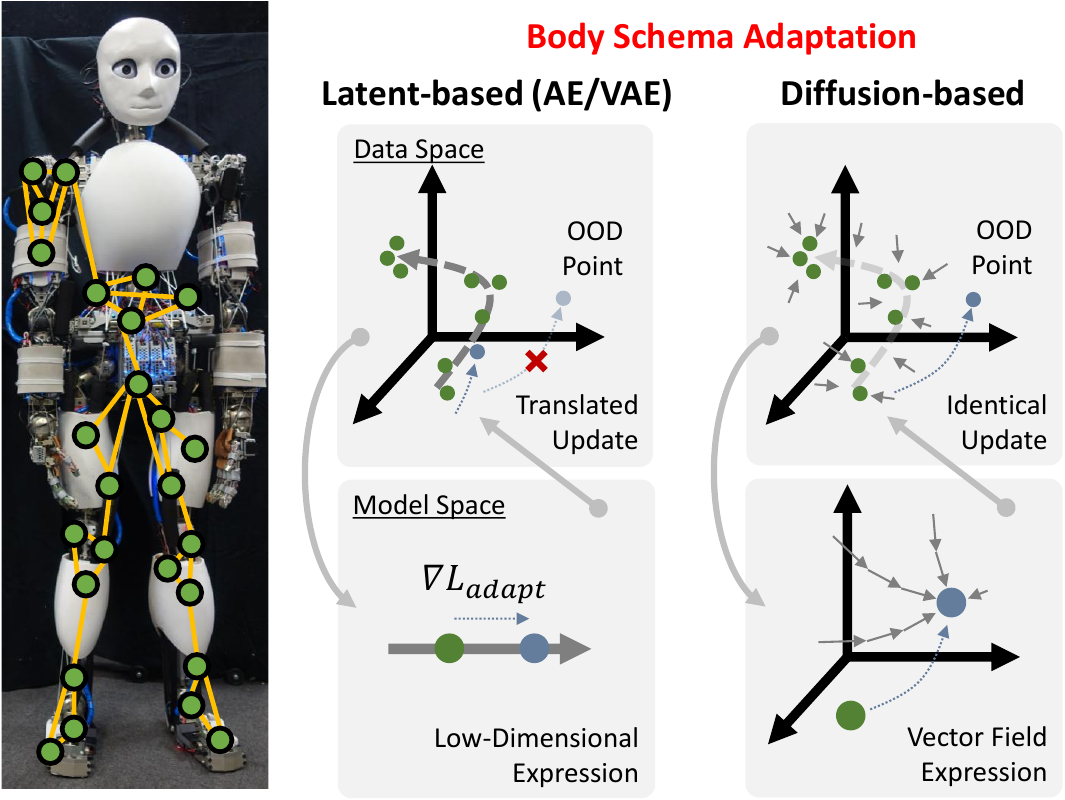}
  \vspace{-1.0ex}
  \caption{Conceptual comparison of musculoskeletal body schema adaptation between low-dimensional latent-based generative models and high-dimensional diffusion-based models. Latent-based models perform updates in a compressed model space, restricting reachable solutions in the data space, whereas diffusion models operate directly in the high-dimensional data space, where all states remain accessible and are shaped through dynamical guidance.}
  \label{figure:concept}
  \vspace{-3.0ex}
\end{figure}

\switchlanguage%
{%
  To fully exploit this adaptability, it is necessary not only to handle redundancy at the hardware level but also to achieve adaptation at the software level, namely at the level of the body schema \cite{gallagher1986bodyschema, haggard2005bodyschema}.
  Accordingly, a wide range of studies has been conducted.
  In general, a body schema realizes appropriate postures by learning the relationships among sensor and actuator variables such as joint angles, muscle lengths, and muscle tensions, and various representations have been employed, including data tables \cite{nakanishi2010estimation}, polynomial approximations \cite{ookubo2015learning}, and neural networks \cite{kawaharazuka2018online}.
  In particular, approaches using neural networks with high representational capacity have become common, and simple regression-based methods \cite{kawaharazuka2019longtime}, autoencoder-based methods \cite{kawaharazuka2020autoencoder}, and variational autoencoder-based methods \cite{yoshimitsu2023vae} have been proposed.
  While simple regression-based methods are easy to train, adapting to muscle abnormalities requires either introducing learning variables that explicitly account for muscle rupture at the network input \cite{tani2002parametric} or retraining the network using data that include abnormal states \cite{kawaharazuka2022redundancy}.
  In contrast, autoencoder-based and variational autoencoder-based methods learn the redundant space of sensor and actuator signals as low-dimensional latent variables, and by manipulating these latent variables, they may adapt to abnormal states without retraining.
  However, because these methods learn average behavior by projecting the redundant space into a low-dimensional latent space, their ability to handle out-of-distribution abnormal states that are not included in the training data is likely to be limited, as illustrated in the left panel of \figref{figure:concept}.

  Accordingly, this study proposes a diffusion-based framework for body schema learning in musculoskeletal robots that enables adaptation to abnormal states.
  Unlike generative models that rely on low-dimensional latent spaces, such as autoencoders and variational autoencoders, diffusion models iteratively estimate physically consistent sensor and actuator values directly in the original high-dimensional space where they are defined, through a denoising process.
  By formulating body schema estimation as a gradient-guided denoising process, we hypothesize that appropriate muscle lengths and muscle tensions can be adaptively estimated even under abnormal conditions such as muscle rupture and actuator jamming.
  Through simulations of a two-joint, six-muscle musculoskeletal robot model, we demonstrate that the proposed diffusion-based body schema can adapt its internal representation and maintain consistent behavior across various actuator failure scenarios without the need for explicit failure annotations during pretraining or any subsequent retraining.
  Furthermore, we validate the effectiveness of the proposed method through comparisons with body schema adaptation approaches based on autoencoders and variational autoencoders.

  This paper is organized as follows.
  \secref{sec:bodyschema} describes the basic structure of musculoskeletal robots and body schema learning methods based on autoencoders, variational autoencoders, and diffusion models.
  \secref{sec:adaptation} presents latent-based and gradient-based body schema adaptation methods for abnormal conditions such as muscle rupture and actuator jamming.
  \secref{sec:exp-setup} details the musculoskeletal robot simulation, body schema learning settings, and evaluation methodologies.
  \secref{sec:experiment} reports comparative experiments of the proposed and baseline body schema learning methods using musculoskeletal robot simulations, followed by discussion.
}%
{%
  この適応能力を引き出すためには, ハードウェアとして冗長性を扱えるだけでなく, ソフトウェア, すなわち身体図式\cite{gallagher1986bodyschema, haggard2005bodyschema}のレベルでも適応が必要であり, これまで様々な研究が行われてきた.
  一般的な身体図式は関節角度・筋長・筋張力などのセンサ・アクチュエータの関係性を学習することで, 適切な姿勢を実現するものであり, データテーブル\cite{nakanishi2010estimation}から多項式近似\cite{ookubo2015learning}, ニューラルネットワーク\cite{kawaharazuka2018online}などの様々な表現が用いられている.
  特に表現能力の高いニューラルネットワークを使った方法が一般的になり, 単純回帰型\cite{kawaharazuka2019longtime}, オートエンコーダ型\cite{kawaharazuka2020autoencoder}, 変分オートエンコーダ型\cite{yoshimitsu2023vae}などが提案されている.
  単純回帰型は学習が容易である一方, 筋の異常に適応するためにはネットワーク入力側に予め筋破断を考慮した学習変数を加える\cite{tani2002parametric}, または異常状態を含む学習データを用いて再学習を行う必要があった\cite{kawaharazuka2022redundancy}.
  これに対して, オートエンコーダ型や変分オートエンコーダ型は, センサ・アクチュエータに関する余剰空間を低次元な潜在変数として学習しており, この潜在変数を操作することで, 再学習無しに異常状態に適応できる可能性がある.
  その一方で, 余剰空間を低次元潜在空間に射影することで平均的な挙動を学習するため, 学習データに含まれない分布外な異常状態への対応能力には限界がある可能性が高い(\figref{figure:concept}の左図).

  そこで本研究では, 筋骨格ロボットにおける異常状態適応可能な身体図式学習に向けた, 拡散モデルに基づく枠組みを提案する.
  拡散モデルは, オートエンコーダや変分オートエンコーダなどの低次元な潜在空間を通した生成モデルとは異なり, ノイズ除去過程に基づき, 再学習なしに, 部分観測や制約条件の下でも物理的整合性の取れたセンサ・アクチュエータ値を反復的に推定する.
  身体図式推定を勾配に基づくガイダンスとノイズ除去の過程として定式化することで, 筋の断裂やアクチュエータの固着といった異常条件下においても, 適切な筋長や筋張力を適応的に推定することができると考えた.
  筋骨格ロボットモデルを用いた2関節6筋シミュレーションを通じて, 提案する拡散モデルに基づく身体図式が, 事前学習における明示的な故障ラベル入力や再学習を必要とすることなく, さまざまなアクチュエータ故障シナリオの下で内部表現を適応させ, 一貫した行動を維持できることを示す.
  また, オートエンコーダや変分オートエンコーダに基づく身体図式適応と比較し, 提案手法の有効性を検証する.

  本研究は以下のように構成される.
  \secref{sec:bodyschema}では, 筋骨格ロボットの基本構造と, オートエンコーダ・変分オートエンコーダ・拡散モデルに基づく身体図式学習手法について述べる.
  \secref{sec:adaptation}では, 筋破断・アクチュエータ固着といった異常状態に適応するための潜在空間, または勾配に基づく身体図式適応手法について述べる.
  \secref{sec:exp-setup}では, 筋骨格ロボットシミュレーションや身体図式の学習設定, 評価方法について述べる.
  \secref{sec:experiment}では, 筋骨格ロボットシミュレーションを用いた各身体図式学習手法の比較実験とその考察を述べる.
}%

\begin{figure}[t]
  \centering
  \includegraphics[width=0.95\columnwidth]{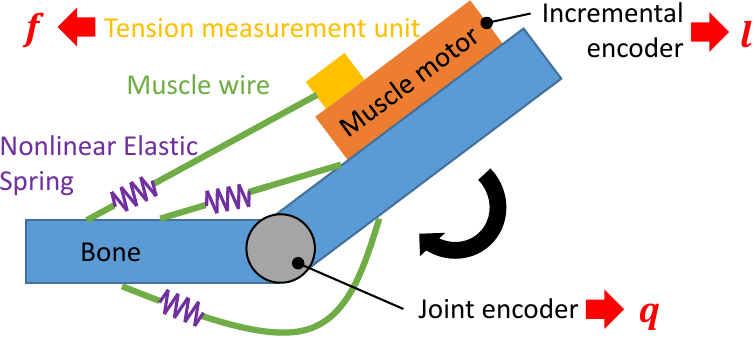}
  \vspace{-1.0ex}
  \caption{The basic structure of musculoskeletal robots used in this study. Muscles are modeled as wires that connect motors to the skeleton via a muscle tension measurement unit and a nonlinear elastic spring at the tip. Muscle length is measured from the motor encoder, and muscle tension is measured from the pressure sensor attached to the muscle tension measurement unit.}
  \label{figure:basic-structure}
  \vspace{-3.0ex}
\end{figure}

\section{Body Schema Learning based on AutoEncoder, Variational AutoEncoder, and Diffusion Model} \label{sec:bodyschema}

\subsection{Basic Structure and Body Schema of Musculoskeletal Robots} \label{subsec:basic}
\switchlanguage%
{%
  We first describe the assumptions of this study.
  As shown in \figref{figure:basic-structure}, a musculoskeletal structure is basically composed of a skeleton, joints, and wires that emulate muscles.
  Depending on the robot, joint angles $\bm{q}$ may be directly available; even when they are not, they can be estimated by combining muscle length changes with visual sensors \cite{kawaharazuka2018online}.
  Each wire is wound around a pulley attached to the motor shaft and is connected to the skeleton through a muscle tension measurement unit.
  In this configuration, the muscle length $l^{motor}$ is obtained from an encoder attached to the motor, while the muscle tension $f$ is obtained from a pressure sensor attached to the muscle tension measurement unit.
  A nonlinear elastic spring is attached to the end of each wire, and this, together with redundancy, enables variable stiffness control.
  The relationship between muscle tension and the extension of the nonlinear elastic spring is modeled as follows:
  \begin{align}
    f_i = k(\exp(\alpha \delta_i) - 1) \label{eq:nonlinear-spring}
  \end{align}
  Here, $k$ denotes the elastic coefficient, $\alpha$ is a constant representing the degree of nonlinearity, and $\delta$ represents the spring extension.

  Next, let the number of joints be $N$ and the number of muscles be $M$, and consider the relationship among the joint angles $\bm{q} \in \mathbb{R}^N$, the muscle lengths $\bm{l}^{motor} \in \mathbb{R}^M$, and the muscle tensions $\bm{f} \in \mathbb{R}^M$.
  For muscle $i$, the following holds:
  \begin{align}
    \delta_i = \phi^{-1}(f_i) = \max(0, l^{geo}_i(\bm{q}) - l^{motor}_i) \label{eq:muscle-elongation}
  \end{align}
  Here, $\bm{l}^{geo}(\bm{q})$ denotes a function representing the geometric muscle lengths that depend on the joint angles.
  When nonlinear elasticity and wire elongation are not considered, the muscle lengths $\bm{l}^{motor}$ obtained from the motors are equal to the geometric muscle lengths.
  In addition, when the muscle tension is positive, the inverse mapping can be computed as follows:
  \begin{align}
    \delta_i = \frac{1}{\alpha}\log(\frac{f_i}{k}+1) \label{eq:muscle-elongation-inv}
  \end{align}

  Finally, we explain the objective of body schema learning in this study.
  The goal of body schema learning in musculoskeletal robots is to learn the relationships among the joint angles $\bm{q}$, muscle lengths $\bm{l}^{motor}$, and muscle tensions $\bm{f}$, so that these quantities can be appropriately estimated.
  In particular, this study aims to estimate the muscle lengths $\bm{l}^{motor}$ and muscle tensions $\bm{f}$ of the robot that realize a given target joint angle $\bm{q}^{ref}$.
  That is, we learn a body schema that takes $\bm{q}$ as input and outputs $\bm{l}^{motor}$ and $\bm{f}$.
  Although evaluation in this study is conducted using the model described in \equref{eq:nonlinear-spring}--\equref{eq:muscle-elongation-inv}, none of this information is available during training or during adaptation to abnormal conditions.
  Using only the learned body schema that captures the relationships among $\bm{q}$, $\bm{l}^{motor}$, and $\bm{f}$, we estimate $\bm{l}^{motor}$ and $\bm{f}$ that achieve $\bm{q}^{ref}$ while adapting to the given abnormal state, regardless of whether muscle length control or muscle tension control is employed.
}%
{%
  本研究の前提条件について説明する.
  まず, 筋骨格構造は\figref{figure:basic-structure}に示すように, 基本的に骨格と関節, 筋肉を模したワイヤから構成される.
  ロボットによっては関節角度$q$が直接得られるが, 得られない場合でも筋長変化と視覚センサを併用することである程度推定可能である\cite{kawaharazuka2018online}.
  ワイヤはモータ先端に取り付いたプーリに巻かれ, 筋張力測定ユニットを通って骨格に接続する.
  このとき, モータに取り付けられたエンコーダから筋長$l^{motor}$が, 筋張力測定ユニットに取り付けられた圧力センサから筋張力$f$が得られる.
  ワイヤの先端には非線形弾性バネが取り付けられ, これが冗長性と合わさって可変剛性制御を可能にする.
  本研究では筋張力と非線形弾性バネの伸びの関係を以下のようにモデル化する.
  \begin{align}
    f_i = k(\exp(\alpha \delta_i) - 1) \label{eq:nonlinear-spring}
  \end{align}
  なお, $k$は弾性係数, $\alpha$は非線形度合いを表す定数, $\delta$はバネの伸びを表す.

  次に, 関節数を$N$, 筋数を$M$とし, 関節角度$\bm{q} \in \mathbb{R}^N$, 筋長$\bm{l}^{motor} \in \mathbb{R}^M$, 筋張力$\bm{f} \in \mathbb{R}^M$の関係を表す.
  筋$i$について, 以下が成り立つ.
  \begin{align}
    \delta_i = \phi^{-1}(f_i) = \max(0, l^{geo}_i(\bm{q}) - l^{motor}_i) \label{eq:muscle-elongation}
  \end{align}
  ここで, $\bm{l}^{geo}(\bm{q})$は関節角度依存の幾何学的な筋長を表す関数である.
  非線形弾性やワイヤの伸びを考慮しない場合, モータから得られる筋長$\bm{l}^{motor}$は幾何学的筋長に等しい.
  また, 張力が正である場合, 逆写像は以下のように計算できる.
  \begin{align}
    \delta_i = \frac{1}{\alpha}\log(\frac{f_i}{k}+1) \label{eq:muscle-elongation-inv}
  \end{align}

  最後に, 本研究における身体図式学習の目的を説明する.
  筋骨格ロボットにおける身体図式学習の目的は, 関節角度$\bm{q}$, 筋長$\bm{l}^{motor}$, 筋張力$\bm{f}$の関係を学習し, これらの値を適切に推定できるようにすることである.
  特に本研究では, 与えられた関節角度$\bm{q}^{ref}$を実現するようなロボットの筋長$\bm{l}^{motor}$と筋張力$\bm{f}$を推定することを目的とする.
  つまり, 入力を$\bm{q}$, 出力を$\bm{l}^{motor}$と$\bm{f}$とする身体図式を学習する.
  なお, 本研究では\equref{eq:nonlinear-spring}--\equref{eq:muscle-elongation-inv}で示されるようなモデルを用いて評価を行うが, 学習時や異常に対する適応時はこれらの情報は一切得られない.
  $\bm{q}$, $\bm{l}^{motor}$, $\bm{f}$の関係性を学習した身体図式のみから, 与えられた異常状態に適応しつつ$\bm{q}^{ref}$を実現する$\bm{l}^{motor}$と$\bm{f}$を推定する(筋長制御であるか筋張力制御であるかは問わない).
}%

\begin{figure*}[t]
  \centering
  \includegraphics[width=1.95\columnwidth]{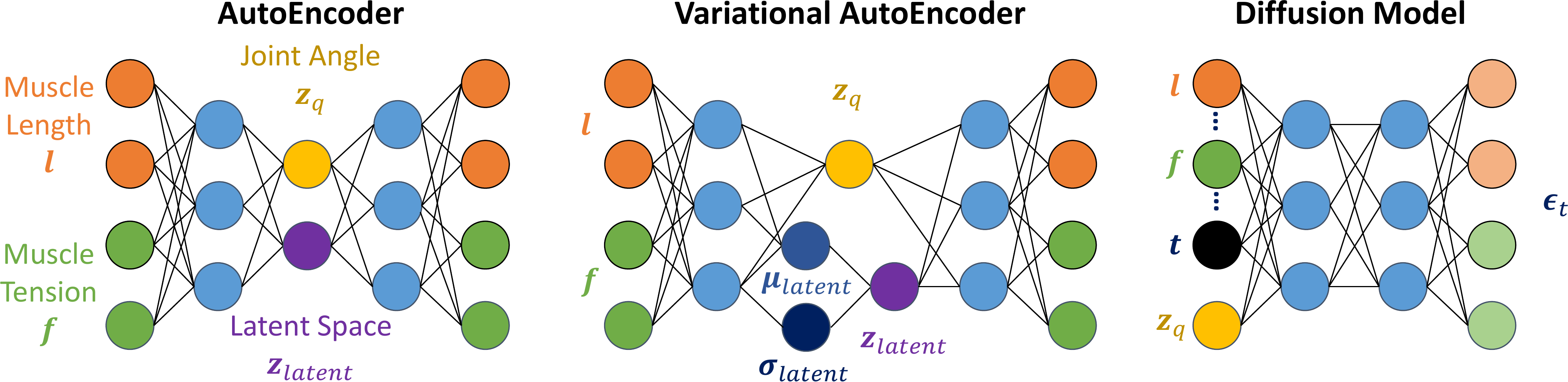}
  \vspace{-1.0ex}
  \caption{Body schema learning methods based on autoencoders (left), variational autoencoders (center), and diffusion models (right).}
  \label{figure:body-schema}
  \vspace{-3.0ex}
\end{figure*}

\subsection{Body Schema Learning based on AutoEncoder and Variational AutoEncoder} \label{subsec:ae-vae}
\switchlanguage%
{%
  First, we describe body schema learning methods based on autoencoders (AE) and variational autoencoders (VAE), as shown in the left and center panels of \figref{figure:body-schema}.
  The autoencoder-based method is constructed as follows:
  \begin{align}
    \bm{x} &= [\bm{l}^{motor}, \bm{f}] \\
    \bm{z}_{q}, \bm{z}_{latent} &= \bm{h}^{ae}_{enc}(\bm{x}) \\
    \bm{\hat{x}} &= \bm{h}^{ae}_{dec}(\bm{z}_{q}, \bm{z}_{latent})
  \end{align}
  Here, $\bm{h}^{ae}_{enc}$ and $\bm{h}^{ae}_{dec}$ denote the encoder and decoder, respectively.
  The latent variable $\bm{z}_{q}$ corresponds to the joint angles $\bm{q}$, while $\bm{z}_{latent}$ represents the latent variables corresponding to the redundant space.
  Note that $\bm{x}$ is normalized using the training data, and hereafter $\bm{x}$ is assumed to be normalized during all training procedures.

  During training, the following loss function is minimized:
  \begin{align}
    L^{ae}_{train} = ||\bm{x} - \bm{\hat{x}}||^2 + C^{train}_{q}||\bm{q} - \bm{z}_{q}||^2
  \end{align}
  Here, $C^{train}_{q}$ is a constant that represents the weight for joint angle estimation.
  By training this autoencoder, muscle lengths $\bm{l}^{motor}$ and muscle tensions $\bm{f}$ can be computed by providing the joint angles $\bm{q}$ and the latent variables $\bm{z}_{latent}$ corresponding to the redundant space as inputs to $\bm{h}^{ae}_{dec}$.

  Next, the variational autoencoder-based method is constructed as follows:
  \begin{align}
    \bm{z}_{q}, \bm{\mu}_{latent}, \bm{\sigma}_{latent} &= \bm{h}^{vae}_{enc}(\bm{x}) \\
    \bm{z}_{latent} &\sim \mathcal{N}(\bm{\mu}_{latent}, \mathrm{diag}(\bm{\sigma}_{latent}^2)) \\
    \bm{\hat{x}} &= \bm{h}^{vae}_{dec}(\bm{z}_{q}, \bm{z}_{latent})
  \end{align}
  Here, $\bm{h}^{vae}_{enc}$ and $\bm{h}^{vae}_{dec}$ denote the encoder and decoder, respectively.

  During training, the following loss function is minimized:
  \begin{align}
    L^{vae}_{train} = &||\bm{x} - \bm{\hat{x}}||^2 + C^{train}_{q}||\bm{q} - \bm{z}_{q}||^2 \nonumber \\
    &+ D_{KL}(\mathcal{N}(\bm{\mu}_{latent}, \mathrm{diag}(\bm{\sigma}_{latent}^2)) || \mathcal{N}(\bm{0}, I))
  \end{align}
  Here, $D_{KL}$ denotes the Kullback-Leibler divergence.
  Similar to the autoencoder-based method, muscle lengths $\bm{l}^{motor}$ and muscle tensions $\bm{f}$ can be computed by providing the joint angles $\bm{q}$ and the latent variables $\bm{z}_{latent}$ corresponding to the redundant space as inputs to $\bm{h}^{vae}_{dec}$.
}%
{%
  まずは, オートエンコーダ(AE)および変分オートエンコーダ(VAE)に基づく身体図式学習手法について説明する(\figref{figure:body-schema}の左図・中図).
  オートエンコーダ型は次のように構成される.
  \begin{align}
    \bm{x} &= [\bm{l}^{motor}, \bm{f}] \\
    \bm{z}_{q}, \bm{z}_{latent} &= \bm{h}^{ae}_{enc}(\bm{x}) \\
    \bm{\hat{x}} &= \bm{h}^{ae}_{dec}(\bm{z}_{q}, \bm{z}_{latent})
  \end{align}
  ここで, $\bm{h}^{ae}_{enc}$はエンコーダ, $\bm{h}^{ae}_{dec}$はデコーダを表す.
  また, $\bm{z}_{q}$は関節角度$\bm{q}$に対応する潜在変数, $\bm{z}_{latent}$は余剰空間に対応する潜在変数を表す.
  なお, $\bm{x}$は学習時のデータを用いて正規化されている(以降全ての学習時に$\bm{x}$は正規化されているとする).

  このとき, 学習時には以下の損失関数を最小化する.
  \begin{align}
    L^{ae}_{train} = ||\bm{x} - \bm{\hat{x}}||^2 + C^{train}_{q}||\bm{q} - \bm{z}_{q}||^2
  \end{align}
  ここで, $C^{train}_{q}$は関節角度推定に対する重みを表す定数である.
  このオートエンコーダを学習することで, 関節角度$\bm{q}$と余剰空間に対応する潜在変数$\bm{z}_{latent}$を$\bm{h}^{ae}_{dec}$に入力することで, 筋長$\bm{l}^{motor}$と筋張力$\bm{f}$を計算できるようになる.

  次に, 変分オートエンコーダ型は次のように構成される.
  \begin{align}
    \bm{z}_{q}, \bm{\mu}_{latent}, \bm{\sigma}_{latent} &= \bm{h}^{vae}_{enc}(\bm{x}) \\
    \bm{z}_{latent} &\sim \mathcal{N}(\bm{\mu}_{latent}, \mathrm{diag}(\bm{\sigma}_{latent}^2)) \\
    \bm{\hat{x}} &= \bm{h}^{vae}_{dec}(\bm{z}_{q}, \bm{z}_{latent})
  \end{align}
  ここで, $\bm{h}^{vae}_{enc}$はエンコーダ, $\bm{h}^{vae}_{dec}$はデコーダを表す.

  このとき, 学習時には以下の損失関数を最小化する.
  \begin{align}
    L^{vae}_{train} = &||\bm{x} - \bm{\hat{x}}||^2 + C^{train}_{q}||\bm{q} - \bm{z}_{q}||^2 \nonumber \\
    &+ D_{KL}(\mathcal{N}(\bm{\mu}_{latent}, \mathrm{diag}(\bm{\sigma}_{latent}^2)) || \mathcal{N}(\bm{0}, I))
  \end{align}
  ここで, $D_{KL}$はカルバック・ライブラー情報量を表す.
  オートエンコーダと同様に, 関節角度$\bm{q}$と余剰空間に対応する潜在変数$\bm{z}_{latent}$を入力することで, 筋長$\bm{l}^{motor}$と筋張力$\bm{f}$を計算できるようになる.
}%

\subsection{Body Schema Learning based on Diffusion Model} \label{subsec:diffusion}
\switchlanguage%
{%
  We describe the body schema learning method based on diffusion models, as shown in the right panel of \figref{figure:body-schema}.
  A diffusion model consists of a noising process and a denoising process.
  First, the noising process is defined as follows:
  \begin{align}
    \bm{x}_t = \sqrt{\bar{\alpha}_t}\bm{x}_0 + \sqrt{1 - \bar{\alpha}_t}\bm{\epsilon}, \quad \bm{\epsilon} \sim \mathcal{N}(\bm{0}, I)
  \end{align}
  Here, $\bm{x}_0 = [\bm{l}^{motor}, \bm{f}]$ represents the sensor and actuator values normalized using the training data, and $\bm{x}_t$ denotes the sensor and actuator values after noise addition at time step $t$.
  The constant $\bar{\alpha}_t$ is determined by a scheduler.

  Next, the denoising process is defined as follows:
  \begin{align}
    \bm{\hat{\epsilon}}_t &= \bm{h}^{diff}(\bm{x}_t, t, \bm{q}) \\
    \bm{x}_{t-1} &= \frac{1}{\sqrt{\alpha_t}}\left(\bm{x}_t - \frac{1 - \alpha_t}{\sqrt{1 - \bar{\alpha}_t}}\bm{\hat{\epsilon}}_t\right) + \sigma_t \bm{\eta}, \quad \bm{\eta} \sim \mathcal{N}(\bm{0}, I)
  \end{align}
  Here, $\bm{h}^{diff}$ denotes the diffusion model, and $\alpha_t$ and $\sigma_t$ are constants determined by the scheduler.
  By training this diffusion model, the noise term $\bm{\hat{\epsilon}}_t$ can be estimated given the joint angles $\bm{q}$ and the noise-corrupted sensor and actuator values $\bm{x}_t$ as inputs.
  By iteratively applying the denoising process, the initial state $\bm{x}_0$ corresponding to $\bm{l}^{motor}$ and $\bm{f}$ can be estimated.
}%
{%
  拡散モデルに基づく身体図式学習手法について説明する(\figref{figure:body-schema}の右図).
  拡散モデルは, ノイズ付加過程とノイズ除去過程から構成される.
  まず, ノイズ付加過程は以下のように定義される.
  \begin{align}
    \bm{x}_t = \sqrt{\bar{\alpha}_t}\bm{x}_0 + \sqrt{1 - \bar{\alpha}_t}\bm{\epsilon}, \quad \bm{\epsilon} \sim \mathcal{N}(\bm{0}, I)
  \end{align}
  ここで, $\bm{x}_0 = [\bm{l}^{motor}, \bm{f}]$は学習時のデータを用いて正規化されたセンサ・アクチュエータ値, $\bm{x}_t$は時刻$t$におけるノイズ付加後のセンサ・アクチュエータ値を表す.
  また, $\bar{\alpha}_t$はスケジューラに基づいて決定される定数である.
  次に, ノイズ除去過程は以下のように定義される.
  \begin{align}
    \bm{\hat{\epsilon}}_t &= \bm{h}^{diff}(\bm{x}_t, t, \bm{q}) \\
    \bm{x}_{t-1} &= \frac{1}{\sqrt{\alpha_t}}\left(\bm{x}_t - \frac{1 - \alpha_t}{\sqrt{1 - \bar{\alpha}_t}}\bm{\hat{\epsilon}}_t\right) + \sigma_t \bm{\eta}, \quad \bm{\eta} \sim \mathcal{N}(\bm{0}, I)
  \end{align}
  ここで, $\bm{h}^{diff}$は拡散モデルを表し, $\alpha_t$と$\sigma_t$はスケジューラに基づいて決定される定数である.
  この拡散モデルを学習することで, 関節角度$\bm{q}$とノイズ付加後のセンサ・アクチュエータ値$\bm{x}_t$を入力することで, ノイズ成分$\bm{\hat{\epsilon}}_t$を計算できるようになる.
  これを用いて, 反復的にノイズ除去を行うことで, $\bm{l}^{motor}$と$\bm{f}$に対応する$\bm{x}_0$を推定できるようになる.
}%

\begin{figure}[t]
  \centering
  \includegraphics[width=0.95\columnwidth]{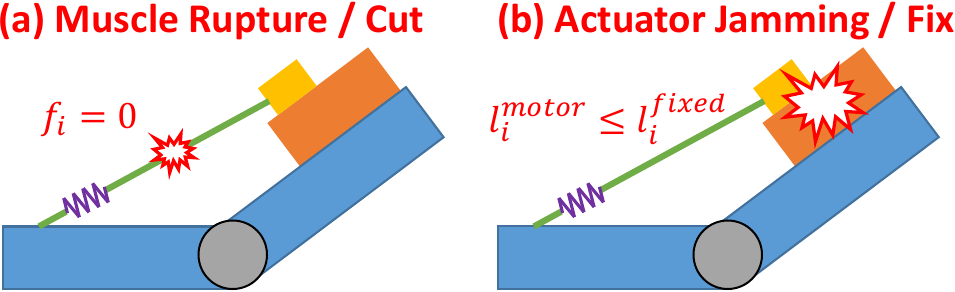}
  \vspace{-1.0ex}
  \caption{Abnormal states considered in this study: (a) muscle rupture (Cut) and (b) actuator jamming (Fix).}
  \label{figure:abnormal-states}
  \vspace{-3.0ex}
\end{figure}

\section{Body Schema Adaptation to Abnormal States} \label{sec:adaptation}
\switchlanguage%
{%
  Here, we describe body schema adaptation methods for coping with abnormal conditions such as muscle rupture and actuator jamming.
  In this study, rather than retraining the model using newly collected data under abnormal conditions, we aim to adapt to such conditions by utilizing a pretrained body schema.
  This approach is important because it enables immediate adaptation when abnormal conditions occur, because appropriate training data for abnormal states may not be available in the first place, and because it helps avoid overfitting caused by retraining.
}%
{%
  ここでは, 筋破断・アクチュエータ固着といった異常状態に適応するための身体図式適応手法について説明する.
  本研究では, 新しく異常状態におけるデータを用いて再学習を行うのではなく, 学習済みの身体図式を用いて異常状態に適応することを目的とする.
  これは, 異常状態が発生した際に即座に適応できること, そもそも異常状態に対する適切な学習データが得られない可能性があること, 再学習によるover-fittingを避けることなどの点で重要である.
}%

\subsection{Definition of Abnormal States} \label{subsec:abnormal}
\switchlanguage%
{%
  We describe the abnormal conditions considered in this study, as illustrated in \figref{figure:abnormal-states}.
  In this study, two types of abnormal conditions are addressed: muscle rupture and actuator jamming.

  First, muscle rupture refers to a state in which a muscle is completely severed and the muscle tension becomes zero.
  When muscle $i$ is ruptured, the following condition holds:
  \begin{align}
    f_i = 0 \label{eq:muscle-cut}
  \end{align}

  Next, actuator jamming refers to a state in which the motor becomes immobile and the muscle length is fixed.
  When muscle $i$ is jammed, the following condition holds:
  \begin{align}
    l^{motor}_i \leq l^{fixed}_i \label{eq:muscle-fix}
  \end{align}
  Here, $l^{fixed}_i$ is a constant representing the muscle length at the moment of jamming.
  This constraint is formulated as an inequality constraint because, due to tension anisotropy of the wire, the muscle length can be shortened but cannot be extended.
}%
{%
  本研究で扱う異常状態について説明する(\figref{figure:abnormal-states}).
  本研究では, 筋破断とアクチュエータ固着の2種類の異常状態を扱う.

  まず, 筋破断は筋が完全に断裂し, 筋張力が0になる状態を表す.
  筋$i$が断裂した場合, 以下の式が成り立つ.
  \begin{align}
    f_i = 0 \label{eq:muscle-cut}
  \end{align}

  次に, アクチュエータ固着はモータが動かなくなり, 筋長が固定される状態を表す.
  筋$i$が固着した場合, 以下の式が成り立つ.
  \begin{align}
    l^{motor}_i \leq l^{fixed}_i \label{eq:muscle-fix}
  \end{align}
  ここで, $l^{fixed}_i$は固着時の筋長を表す定数である.
  この制約は, ワイヤの張力異方性により, 筋長を短くすることはできても, 引き伸ばすことはできないため, 不等式制約として定義される.
}%

\subsection{Body Schema Adaptation based on Latent Space Optimization} \label{subsec:ae-vae-adapt}
\switchlanguage%
{%
  We describe body schema adaptation methods based on autoencoders and variational autoencoders for adapting to the abnormal conditions defined in \secref{subsec:abnormal}.
  Adaptation of these body schemas is performed via latent space optimization.
  Specifically, the latent variable $\bm{z}_{latent}$ is updated as follows.
  \begin{align}
    \bm{\hat{x}} = &\bm{h}^{\{ae, vae\}}_{dec}(\bm{q}, \bm{z}_{latent}) \label{eq:ae-vae}\\
    L_{adapt}(\bm{\hat{x}}) = &C^{adapt}_{cut}\sum_{i \in \mathcal{I}_{cut}} ||f_i||^2 \nonumber \\
    &+ C^{adapt}_{fix}\sum_{j \in \mathcal{I}_{fix}} ||\max(0, l^{motor}_j - l^{fixed}_j)||^2 \nonumber \\
    &+ C^{adapt}_{fmin} \sum_{i=1}^{M} \max(0, -f_i)^2 \nonumber \\
    &+ C^{adapt}_{fmax} \sum_{i=1}^{M} \max(0, f_i - f^{max}_i)^2 \nonumber \\
    \bm{z}_{latent} \leftarrow &\bm{z}_{latent} - C^{adapt}_{lr}\nabla_{\bm{z}_{latent}} L_{adapt}(\bm{\hat{x}}) \label{eq:latent-opt}
  \end{align}
  Here, $\mathcal{I}_{cut}$ denotes the index set of ruptured muscles, $\mathcal{I}_{fix}$ denotes the index set of jammed muscles, and $f^{max}$ is a constant representing the maximum muscle tension.
  In addition, $C^{adapt}_{cut}$, $C^{adapt}_{fix}$, $C^{adapt}_{fmin}$, and $C^{adapt}_{fmax}$ are constants representing the weights for each evaluation term, and $C^{adapt}_{lr}$ is a constant representing the learning rate.
  In other words, adaptation to abnormal conditions is achieved by updating the latent variable $\bm{z}_{latent}$ so as to satisfy the constraints of muscle rupture and actuator jamming, while ensuring that the muscle tension remains nonnegative and does not exceed the maximum muscle tension.
}%
{%
  \secref{subsec:abnormal}で定義した異常状態に適応するための, オートエンコーダおよび変分オートエンコーダに基づく身体図式適応手法について説明する.
  これらの身体図式の適応は, 潜在空間最適化に基づいて行われる.
  具体的には, 以下のように潜在変数$\bm{z}_{latent}$を更新する.
  \begin{align}
    \bm{\hat{x}} = &\bm{h}^{\{ae, vae\}}_{dec}(\bm{q}, \bm{z}_{latent}) \\
    L_{adapt}(\bm{\hat{x}}) = &C^{adapt}_{cut}\sum_{i \in \mathcal{I}_{cut}} ||f_i||^2 \nonumber \\
    &+ C^{adapt}_{fix}\sum_{j \in \mathcal{I}_{fix}} ||\max(0, l^{motor}_j - l^{fixed}_j)||^2 \nonumber \\
    &+ C^{adapt}_{fmin} \sum_{i=1}^{M} \max(0, -f_i)^2 \nonumber \\
    &+ C^{adapt}_{fmax} \sum_{i=1}^{M} \max(0, f_i - f^{max}_i)^2 \nonumber \\
    \bm{z}_{latent} \leftarrow &\bm{z}_{latent} - C^{adapt}_{lr}\nabla_{\bm{z}_{latent}} L_{adapt}(\bm{\hat{x}}) \label{eq:latent-opt}
  \end{align}
  ここで, $\mathcal{I}_{cut}$は断裂した筋のインデックス集合, $\mathcal{I}_{fix}$は固着した筋のインデックス集合, $f^{max}$は筋の最大筋張力を表す定数である.
  また, $C^{adapt}_{cut}$, $C^{adapt}_{fix}$, $C^{adapt}_{fmin}$, $C^{adapt}_{fmax}$はそれぞれの評価項目に対する重みを表す定数, $C^{adapt}_{lr}$は学習率を表す定数である.
  つまり, 筋破断・アクチュエータ固着の制約を満たしつつ, 筋張力が負にならないように, かつ最大筋張力を超えないように潜在変数$\bm{z}_{latent}$を更新することで, 異常状態に適応する.
}%

\subsection{Body Schema Adaptation based on Gradient-Guided Denoising Process} \label{subsec:diff-adapt}
\switchlanguage%
{%
  We describe the body schema adaptation method based on diffusion models for adapting to the abnormal conditions defined in \secref{subsec:abnormal}.
  Diffusion-based body schema adaptation is performed by combining gradient-based guidance with the denoising process.
  Specifically, the denoising process is modified as follows:
  \begin{align}
    \bm{\hat{x}}_0 &= \frac{\bm{x}_t - \sqrt{1-\bar{\alpha}_t}\bm{\hat{\epsilon}}}{\sqrt{\bar{\alpha}_t}} \label{eq:diff-1}\\
    \bm{x}_{t} &\leftarrow \bm{x}_t - C^{adapt}_{lr} \nabla_{\bm{x}_t} L_{adapt}(\bm{\hat{x}}_0) \label{eq:diff-adapt}\\
    \bm{x}_{t-1} &= \frac{1}{\sqrt{\alpha_t}}\left(\bm{x}_t - \frac{1 - \alpha_t}{\sqrt{1 - \bar{\alpha}_t}}\bm{\hat{\epsilon}}_t\right) \label{eq:diff-2}
  \end{align}
  Here, $L_{adapt}(\bm{\hat{x}}_0)$ uses the loss function defined in \equref{eq:latent-opt}.
  In this process, $\bm{\eta}$ is set to zero, and no additional noise is injected.
}%
{%
  \secref{subsec:abnormal}で定義した異常状態に適応するための, 拡散モデルに基づく身体図式適応手法について説明する.
  拡散モデル型の身体図式適応は, 勾配に基づくガイダンスとノイズ除去過程の組み合わせに基づいて行われる.
  具体的には, 以下のようにノイズ除去過程を変更する.
  \begin{align}
    \bm{\hat{x}}_0 &= \frac{\bm{x}_t - \sqrt{1-\bar{\alpha}_t}\bm{\hat{\epsilon}}}{\sqrt{\bar{\alpha}_t}}\\
    \bm{x}_{t} &\leftarrow \bm{x}_t - C^{adapt}_{lr} \nabla_{\bm{x}_t} L_{adapt}(\bm{\hat{x}}_0) \\
    \bm{x}_{t-1} &= \frac{1}{\sqrt{\alpha_t}}\left(\bm{x}_t - \frac{1 - \alpha_t}{\sqrt{1 - \bar{\alpha}_t}}\bm{\hat{\epsilon}}_t\right)
  \end{align}
  ここで, $L_{adapt}(\bm{\hat{x}}_0)$は\equref{eq:latent-opt}で定義される損失関数を用いる.
  この際, $\bm{\eta}$は0に設定し, ノイズ成分を加えない.
}%

\begin{figure}[t]
  \centering
  \includegraphics[width=0.7\columnwidth]{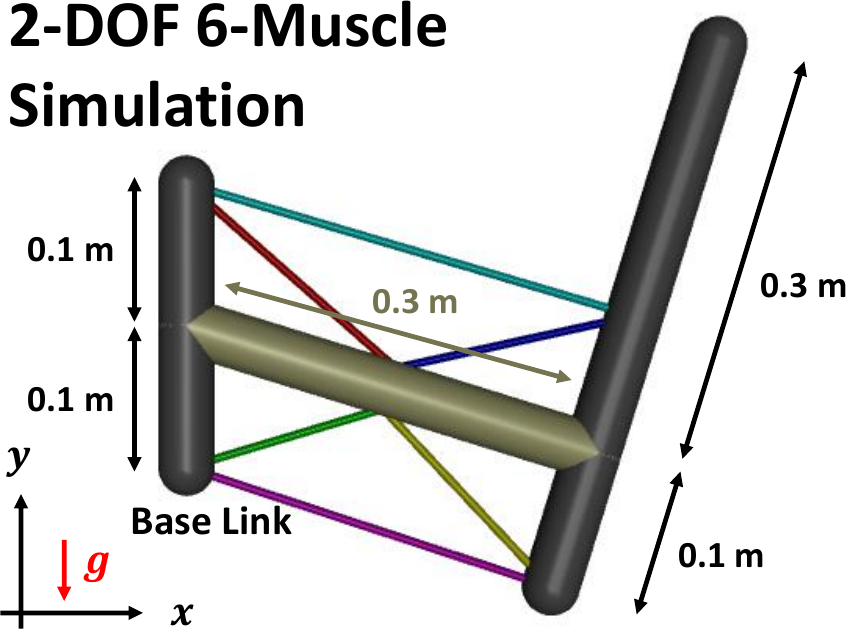}
  \vspace{-1.0ex}
  \caption{Experimental setup using a two-joint, six-muscle musculoskeletal model.}
  \label{figure:exp-setup}
  \vspace{-3.0ex}
\end{figure}

\section{Experimental Setup} \label{sec:exp-setup}

\switchlanguage%
{%
  The simulation environment using the two-joint, six-muscle model considered in this study is shown in \figref{figure:exp-setup}.
  From a base link of length 0.2 m, two links of lengths 0.3 m and 0.4 m, each with a mass of 2.0 kg, are serially connected by rotational joints, and four monoarticular muscles and two biarticular muscles are attached.
  Each joint is movable within the range of [-1.0, 1.0] rad, and gravity acts in the negative direction of the $y$-axis.
  The parameters are set to $f^{max}=150$ N, $k=1000$ N, and $\alpha=20.0$ m$^{-1}$.
}%
{%
  本研究で扱う2関節6筋モデルを用いたシミュレーション環境を\figref{figure:exp-setup}に示す.
  長さ0.2 mのBase Linkから, 長さ0.3 mと0.4 mの重さ各2.0 kgのリンクが回転関節により直列に接続されており, 4本の単関節筋と2本の二関節筋が取り付けられている.
  各関節は[-1.0, 1.0] radの範囲で可動し, $y$軸のマイナス方向に重力が働いている.
  また, $f^{max}=150$ N, $k=1000$ N, $\alpha=20.0$ m$^{-1}$と設定している.
}%

\subsection{Data Collection and Training of Body Schema} \label{subsec:collect-train}
\switchlanguage%
{%
  We describe the data collection procedure for body schema learning.
  In data collection, $N^{data}$ samples of $(\bm{q}, \bm{f}, \bm{l}^{motor})$ are generated.
  The joint angles $\bm{q}$ are uniformly sampled, and the geometric muscle lengths $\bm{l}^{geo}(\bm{q})$ and their derivatives, namely the muscle Jacobian $\bm{G}(\bm{q})$, are obtained using forward kinematics in MuJoCo.
  Then, muscle tensions $\bm{f}$ that satisfy static equilibrium are computed by solving the following optimization problem:
  \begin{align}
    \arg\min_{\bm{f}\in\mathbb{R}^M}\quad & ||\bm{f} - \bm{f}^{ref}||^2_2 \\
    \text{s.t.}\quad & -\bm{G}(\bm{q})^\top \bm{f} = \bm{\tau}^{bias}(\bm{q}) \notag\\
    & \bm{0} \leq \bm{f} \leq \bm{f}^{\max} \notag
  \end{align}
  Here, $\bm{f}^{ref}$ denotes a random vector sampled from the range $[\bm{0}, \bm{f}^{max}]$, and $\bm{\tau}^{bias}(\bm{q})$ represents the torque that the robot should generate, which in this study corresponds to the gravity compensation torque.
  Finally, the muscle lengths $\bm{l}^{motor}$ are computed from \equref{eq:muscle-elongation}.
  In this way, a dataset with diverse muscle tension patterns that satisfy the required torques is generated and used for training.

  Next, we describe the model architectures of the three body schema models: autoencoder, variational autoencoder, and diffusion model.
  The autoencoder and variational autoencoder consist of encoders and decoders implemented as three-layer multilayer perceptrons (MLPs), where each hidden layer has 64 units and the activation function is Tanh.
  For these models, comparisons are conducted with latent space dimensions of 3, 10, and 30, resulting in approximately 10k, 11k, and 14k total parameters, respectively, and approximately 11k, 12k, and 15k parameters in the case of the variational autoencoder.
  Note that the latent space dimensionality is the sum of the dimensions of $z_{q}$ and $z_{latent}$.
  Since the dimensionality of $z_{q}$ is 2 in this study, the dimensionality of $z_{latent}$ becomes 1, 8, and 28, respectively.
  The diffusion model is implemented as a three-layer MLP with two hidden layers of 91 units each, uses the SiLU activation function, and has approximately 11k parameters.
  Across all models, the number of parameters is kept as similar as possible.

  These three models are trained using $N^{data}=800$, with 80 percent of the data used for training and 20 percent for evaluation, over 10000 epochs.
  For the diffusion model, the number of diffusion steps is set to 200, and a linear scheduler is used.
  Other parameters are set to $C^{train}_{q}=2.0$, $C^{adapt}_{cut}=0.01$, $C^{adapt}_{fix}=1.0$, $C^{adapt}_{fmin}=0.01$, $C^{adapt}_{fmax}=0.01$, and $C^{adapt}_{lr}=0.01$.
}%
{%
  身体図式学習のためのデータ収集方法について述べる.
  データ収集では, $(\bm{q}, \bm{f}, \bm{l}^{motor})$のデータを$N^{data}$個生成する.
  関節角度$\bm{q}$を一様にサンプリングし, MuJoCoの順運動学を用いて幾何学的筋長$\bm{l}^{geo}(\bm{q})$とその微分である筋長ヤコビアン$\bm{G}(\bm{q})$を取得する.
  その後, 以下のように静力学的な釣り合いを満たす筋張力$\bm{f}$を計算する.
  \begin{align}
    \arg\min_{\bm{f}\in\mathbb{R}^M}\quad & ||\bm{f} - \bm{f}^{ref}||^2_2 \\
    \text{s.t.}\quad & -\bm{G}(\bm{q})^\top \bm{f} = \bm{\tau}^{bias}(\bm{q}) \notag\\
    & \bm{0} \leq \bm{f} \leq \bm{f}^{\max} \notag
  \end{align}
  ここで, $\bm{f}^{ref}$は$[\bm{0}, \bm{f}^{max}]$までのランダムな値, $\bm{\tau}^{bias}(\bm{q})$はロボットが発揮するべきトルク, ここでは特に重力補償トルクを表す.
  最後に, \equref{eq:muscle-elongation}から筋長$\bm{l}^{motor}$を計算する.
  つまり, 発揮すべきトルクを満たしつつ, 様々な筋張力パターンを持つデータセットを生成, これを学習に利用する.

  AutoEncoder, Variational AutoEncoder, Diffusion Modelの3つの身体図式のモデル構成について述べる.
  AutoEncoderおよびVariational AutoEncoderは, 3層のMLPで構成されたエンコーダとデコーダを持ち, 各中間層のユニット数は64, 活性化関数はTanhである.
  これらは潜在空間のユニット数を3, 10, 30として比較を行い, それぞれの場合で合計のパラメータ数は約10k, 11k, 14kである(VAEの場合は約11k, 12k, 15kである).
  なお, 潜在空間の次元は$z_{q}$と$z_{latent}$の次元の和であり, 本研究では$z_{q}$の次元が2であるため, $z_{latent}$の次元は1, 8, 28となる.
  Diffusion Modelは, 3層のMLPで構成され, 2つの中間層のユニット数は91, 活性化関数SiLUであり, パラメータ数は約11kである.
  各モデルにおいて, そのパラメータ数はなるべく揃えるようにしている.
  これら3つのモデルを, $N^{data}=800$として, その8割を学習データ, 2割を評価データにし, 10000エポックの学習を行う.
  なお, 拡散モデルについては, 拡散ステップ数を200とし, 線形スケジューラを用いている.
  その他のパラメータについては, $C^{train}_{q}=2.0$, $C^{adapt}_{cut}=0.01$, $C^{adapt}_{fix}=1.0$, $C^{adapt}_{fmin}=0.01$, $C^{adapt}_{fmax}=0.01$, $C^{adapt}_{lr}=0.01$に設定している.
}%

\subsection{Evaluation} \label{subsec:exp-eval}
\switchlanguage%
{%
  We describe the evaluation methodology for body schema learning.
  Here, we evaluate whether the muscle lengths $\bm{l}^{motor}$ and muscle tensions $\bm{f}$ computed in \secref{sec:adaptation} can satisfy the conditions in \equref{eq:muscle-cut} and \equref{eq:muscle-fix}, achieve joint torque equilibrium, and realize the target joint angles $\bm{q}^{ref}$.
  An abnormal condition in which two randomly selected muscles out of six are ruptured is defined as \textbf{CUT}, an abnormal condition in which two muscles are jammed is defined as \textbf{FIX}, and an abnormal condition in which one muscle is ruptured and one muscle is jammed is defined as \textbf{BOTH}.
  Adaptation is performed for each abnormal condition.
  The adaptation performance of each model is evaluated using 40 sampled abnormal cases.

  For evaluation, the joint angles $\bm{\hat{q}}$ that would be realized given $\bm{l}^{motor}$ and $\bm{f}$ are computed using forward kinematics in MuJoCo, static equilibrium, and the abnormal condition constraints.
  This is achieved by iteratively updating $\bm{\hat{q}}$ using the Gauss-Newton method under multiple constraints.
  Since physical simulation including muscle slack is difficult, an optimization-based approach is adopted.
  First, forward kinematics in MuJoCo is used to obtain the geometric muscle lengths $\bm{l}^{geo}(\bm{\hat{q}})$, the muscle Jacobian $\bm{G}(\bm{\hat{q}})$, and bias torques such as gravity compensation torque $\bm{\tau}^{bias}(\bm{\hat{q}})$.
  Here, the set of taut muscles $\mathcal{E}(\bm{\hat{q}})$ is defined based on a tension threshold and geometric conditions as follows:
  \begin{align}
    \mathcal{E}(\bm{\hat{q}})
    &= \Bigl\{i \notin \bigl(\mathcal{I}_{cut}\cup \mathcal{I}_{fix}\bigr) \Bigm| f_i>0, \notag\\
    &\hspace{2.2em} l_i^{geo}(\bm{\hat{q}})-l_i^{motor} \ge -\varepsilon^{slack} \Bigr\}
  \end{align}
  Here, $\varepsilon^{slack}$ is a constant representing the slack tolerance.
  For taut muscles $i\in\mathcal{E}(\bm{\hat{q}})$, equality residuals are defined using the elongation $\delta_i=\phi^{-1}(f_i)$:
  \begin{align}
    r_i^{eq}(\bm{\hat{q}})
    &=\left(l_i^{motor}+\phi^{-1}(f_i)\right)-l_i^{geo}(\bm{\hat{q}}), \notag\\
    &\qquad i\in\mathcal{E}(\bm{\hat{q}})
  \end{align}
  For slack muscles $i\notin\mathcal{E}(\bm{\hat{q}})$, the violation of the slack condition is introduced as a hinge residual:
  \begin{align}
    r_i^{slack}(\bm{\hat{q}})
    &= \max\!\left(0,\ l_i^{geo}(\bm{\hat{q}})-l_i^{motor}\right), \notag\\
    &\qquad i\notin\mathcal{E}(\bm{\hat{q}}), i\notin \mathcal{I}_{cut}\cup\mathcal{I}_{fix}
  \end{align}
  For jammed muscles $i\in\mathcal{I}_{fix}$, a one-sided constraint is imposed using the fixed motor length $l_i^{fix}$:
  \begin{align}
    r_i^{fix}(\bm{\hat{q}})
    &=\max\left(0,\ l_i^{geo}(\bm{\hat{q}})-\left(l_i^{fix}+\phi^{-1}(f_i)\right)\right), \notag\\
    & \qquad i\in\mathcal{I}_{fix}
  \end{align}
  These residuals are solved as an Iteratively Reweighted Least Squares (IRLS) problem with weights $w_i$ assigned to each residual, minimizing the Huber loss \cite{holland1977robust}.

  In addition, a static equilibrium regularization term is introduced.
  We define the effective muscle tension $\bm{f}^{eff}$ using only taut muscles as
  \begin{align}
    f_i^{eff}=
    \begin{cases}
      f_i & (i\in\mathcal{E}(\bm{\hat{q}}))\\
      0 & \text{otherwise}
    \end{cases}
  \end{align}
  Using this definition, the joint torque imbalance between the muscle-generated torque and the bias torque is given by
  \begin{align}
    \bm{\tau}(\bm{\hat{q}}) = -\bm{G}(\bm{\hat{q}})^\top \bm{f}^{eff} - \bm{\tau}^{bias}(\bm{\hat{q}})
  \end{align}
  and is added as a regularization term.

  To compensate for differences in residual units, a length scale $s_l$ and a torque scale $s_\tau$ are introduced.
  Finally, $\bm{\hat{q}}$ is estimated by solving the following minimization problem:
  \begin{align}
    \min_{\bm{\hat{q}}\in\mathbb{R}^2}\
    \sum_i w_i\left(\frac{r_i(\bm{\hat{q}})}{s_l}\right)^2
    +\lambda_{\tau}\left|\left|\frac{\bm{\tau}(\bm{\hat{q}})}{s_\tau}\right|\right|_2^2
    +\lambda_q||\bm{\hat{q}}-\bm{q}^{ref}||_2^2
  \end{align}
  Here, $r_i(\bm{\hat{q}})$ is chosen from $\{r_i^{eq}, r_i^{slack}, r_i^{fix}\}$ depending on the muscle state, and ruptured muscles $i\in\mathcal{I}_{cut}$ are completely excluded.
  The constant $\lambda_{\tau}$ represents the weight of the static equilibrium regularization, and $\lambda_q$ represents the weight of the joint angle regularization.
  The residual vector $\bm{r}(\bm{\hat{q}})$ is linearized as follows, and $\bm{\hat{q}}$ is iteratively updated using the Gauss-Newton method:
  \begin{align}
    \bm{r}(\bm{\hat{q}}+\Delta\bm{\hat{q}}) &\approx \bm{r}(\bm{\hat{q}}) + \bm{J}(\bm{\hat{q}})\Delta\bm{\hat{q}}\\
    \left(\bm{J}^\top\bm{J}+\mu\bm{I}\right)\Delta\bm{\hat{q}} &= -\bm{J}^\top \bm{r}\\
    \bm{\hat{q}} &\leftarrow \bm{\hat{q}}+\Delta\bm{\hat{q}}
  \end{align}
  Through this estimation, given the input $(\bm{l}^{motor},\bm{f})$, the joint angles $\bm{\hat{q}}$ that simultaneously satisfy rupture and jamming conditions, slack constraints, and static equilibrium are reconstructed.

  Next, we define four evaluation metrics, which represent the tension error under muscle rupture, the muscle length error under actuator jamming, the static equilibrium error, and the joint angle error, respectively.
  \begin{align}
    E_{cut} &= \frac{1}{|\mathcal{I}_{cut}|}\sum_{i \in \mathcal{I}_{cut}} |f_i| \\
    E_{fix} &= \frac{1}{|\mathcal{I}_{fix}|}\sum_{j \in \mathcal{I}_{fix}} \max(0, l^{motor}_j - l^{fixed}_j) \\
    E_{\tau} &= ||-\bm{G}(\bm{\hat{q}})^\top \bm{f}^{eff} - \bm{\tau}^{bias}(\bm{\hat{q}})||_2 \\
    E_{q}   &= ||\bm{q}^{ref} - \bm{\hat{q}}||_2
  \end{align}
  In this study, the parameters are set to $\varepsilon^{slack}=10^{-5}$ m, $s_l=0.03$ m, $s_{\tau}=5.0$ Nm, $\lambda_{\tau}=0.1$, and $\lambda_q=10^{-3}$.
}%
{%
  身体図式学習の評価方法について述べる.
  ここでは, \secref{sec:adaptation}で計算された筋長$\bm{l}^{motor}$と筋張力$\bm{f}$が, \equref{eq:muscle-cut}や\equref{eq:muscle-fix}の条件, 関節トルクの釣り合い, 目標関節角度$\bm{q}^{ref}$を実現できるかどうかを評価する.
  6本中ランダムな2本の筋肉が破断している異常状態を\textbf{CUT}, 2本の筋肉が固着している異常状態を\textbf{FIX}, 1本の筋肉が破断し1本の筋肉が固着している異常状態を\textbf{BOTH}と定義し, 各異常状態に対して適応を行う.
  サンプリングされた40の異常サンプルを用いて, 各モデルの適応性能を評価する.

  評価に必要なため, $\bm{l}^{motor}$と$\bm{f}$が与えられた際に実現されるであろう$\bm{\hat{q}}$を, MuJoCoの順運動学と静力学的釣り合い, 異常状態の条件を用いて求める.
  これは, $\bm{\hat{q}}$を複数の制約条件からGauss-Newton法で反復更新することで実現する(筋の緩みを含めた物理シミュレーションが難しいため, 最適化を用いた).
  まずは, MuJoCoの順運動学により$\bm{l}^{geo}(\bm{\hat{q}})$と筋長ヤコビアン$\bm{G}(\bm{\hat{q}})$, および重力補償トルク等のバイアストルク$\bm{\tau}^{bias}(\bm{\hat{q}})$を得る.
  ここで, 張っている筋の集合$\mathcal{E}(\bm{\hat{q}})$を, 張力閾値と幾何条件から定義する.
  \begin{align}
    \mathcal{E}(\bm{\hat{q}})
    &= \Bigl\{i \notin \bigl(\mathcal{I}_{cut}\cup \mathcal{I}_{fix}\bigr) \Bigm| f_i>0, \notag\\
    &\hspace{2.2em} l_i^{geo}(\bm{\hat{q}})-l_i^{motor} \ge -\varepsilon^{slack} \Bigr\}
  \end{align}
  ここで, $\varepsilon^{slack}$は緩み許容値を表す定数である.
  張っている筋$i\in\mathcal{E}(\bm{\hat{q}})$について, 伸び$\delta_i=\phi^{-1}(f_i)$を用いて, 等式残差を定義する.
  \begin{align}
    r_i^{eq}(\bm{\hat{q}})
    &=\left(l_i^{motor}+\phi^{-1}(f_i)\right)-l_i^{geo}(\bm{\hat{q}}), \notag\\
    &\qquad i\in\mathcal{E}(\bm{\hat{q}})
  \end{align}
  張っていない筋$i\notin\mathcal{E}(\bm{\hat{q}})$について, 緩み条件の違反量をヒンジ残差として導入する.
  \begin{align}
    r_i^{slack}(\bm{\hat{q}})
    &= \max\!\left(0,\ l_i^{geo}(\bm{\hat{q}})-l_i^{motor}\right), \notag\\
    &\qquad i\notin\mathcal{E}(\bm{\hat{q}}), i\notin \mathcal{I}_{cut}\cup\mathcal{I}_{fix}
  \end{align}
  焼き付き筋$i\in\mathcal{I}_{fix}$について, 固定モータ長$l_i^{fix}$を用いて片側拘束を課す.
  \begin{align}
    r_i^{fix}(\bm{\hat{q}})
    &=\max\left(0,\ l_i^{geo}(\bm{\hat{q}})-\left(l_i^{fix}+\phi^{-1}(f_i)\right)\right), \notag\\
    & \qquad i\in\mathcal{I}_{fix}
  \end{align}
  これらはHuber損失を最小化する, 各残差に重み$w_i$を付けたIteratively Reweighted Least Squares (IRLS)として解く\cite{holland1977robust}.

  これに加えて, 静力学バランス正則化項も加える.
  張っている筋のみを用いた有効張力$\bm{f}^{eff}$を定義する.
  \begin{align}
    f_i^{eff}=
    \begin{cases}
      f_i & (i\in\mathcal{E}(\bm{\hat{q}}))\\
      0 & \text{otherwise}
    \end{cases}
  \end{align}
  このとき, 筋張力による関節トルクとバイアストルクの釣り合い誤差を
  \begin{align}
    \bm{\tau}(\bm{\hat{q}}) = -\bm{G}(\bm{\hat{q}})^\top \bm{f}^{eff} - \bm{\tau}^{bias}(\bm{\hat{q}})
  \end{align}
  として正則化項に加える.

  残差の単位差を補正するため, 長さスケール$s_l$とトルクスケール$s_\tau$を導入する.
  最終的に$\bm{\hat{q}}$は次の最小化で推定される.
  \begin{align}
    \min_{\bm{\hat{q}}\in\mathbb{R}^2}\
    \sum_i w_i\left(\frac{r_i(\bm{\hat{q}})}{s_l}\right)^2
    +\lambda_{\tau}\left|\left|\frac{\bm{\tau}(\bm{\hat{q}})}{s_\tau}\right|\right|_2^2
    +\lambda_q||\bm{\hat{q}}-\bm{q}^{ref}||_2^2
  \end{align}
  ここで$r_i(\bm{\hat{q}})$は, 筋の状態に応じて$\{r_i^{eq}, r_i^{slack}, r_i^{fix}\}$のいずれかを用いる(切断された筋$i\in\mathcal{I}_{cut}$は完全に除外する).
  また, $\lambda_{\tau}$は静力学バランス正則化の重みを表す定数, $\lambda_q$は関節角度正則化の重みを表す定数である.
  以下のように残差ベクトル$\bm{r}(\bm{\hat{q}})$を線形化し, Gauss-Newton法により反復的に更新する.
  \begin{align}
    \bm{r}(\bm{\hat{q}}+\Delta\bm{\hat{q}}) &\approx \bm{r}(\bm{\hat{q}}) + \bm{J}(\bm{\hat{q}})\Delta\bm{\hat{q}}\\
    \left(\bm{J}^\top\bm{J}+\mu\bm{I}\right)\Delta\bm{\hat{q}} &= -\bm{J}^\top \bm{r}\\
    \bm{\hat{q}} &\leftarrow \bm{\hat{q}}+\Delta\bm{\hat{q}}
  \end{align}
  この推定により, 入力 $(\bm{l}^{motor},\bm{f})$が与えられたときに, 切断・固着条件, 緩み制約, 静力学バランスを同時に満たす$\bm{q}$を復元する.

  ここで, 4つの評価指標を示す.
  それぞれ, 筋切断の張力誤差, 筋固着の筋長誤差, 静力学バランス誤差, 関節角度誤差を表している.
  \begin{align}
    E_{cut} &= \frac{1}{|\mathcal{I}_{cut}|}\sum_{i \in \mathcal{I}_{cut}} |f_i| \\
    E_{fix} &= \frac{1}{|\mathcal{I}_{fix}|}\sum_{j \in \mathcal{I}_{fix}} \max(0, l^{motor}_j - l^{fixed}_j) \\
    E_{\tau} &= ||-\bm{G}(\bm{\hat{q}})^\top \bm{f}^{eff} - \bm{\tau}^{bias}(\bm{\hat{q}})||_2 \\
    E_{q}   &= ||\bm{q}^{ref} - \bm{\hat{q}}||_2
  \end{align}
  なお, 本研究では$\varepsilon^{slack}=1.0e^{-5}$ m, $s_l=0.03$ m, $s_{\tau}=5.0$ Nm, $\lambda_{\tau}=0.1$, $\lambda_q=1.0e^{-3}$と設定している.
}%

\begin{figure*}[t]
  \centering
  \includegraphics[width=1.9\columnwidth]{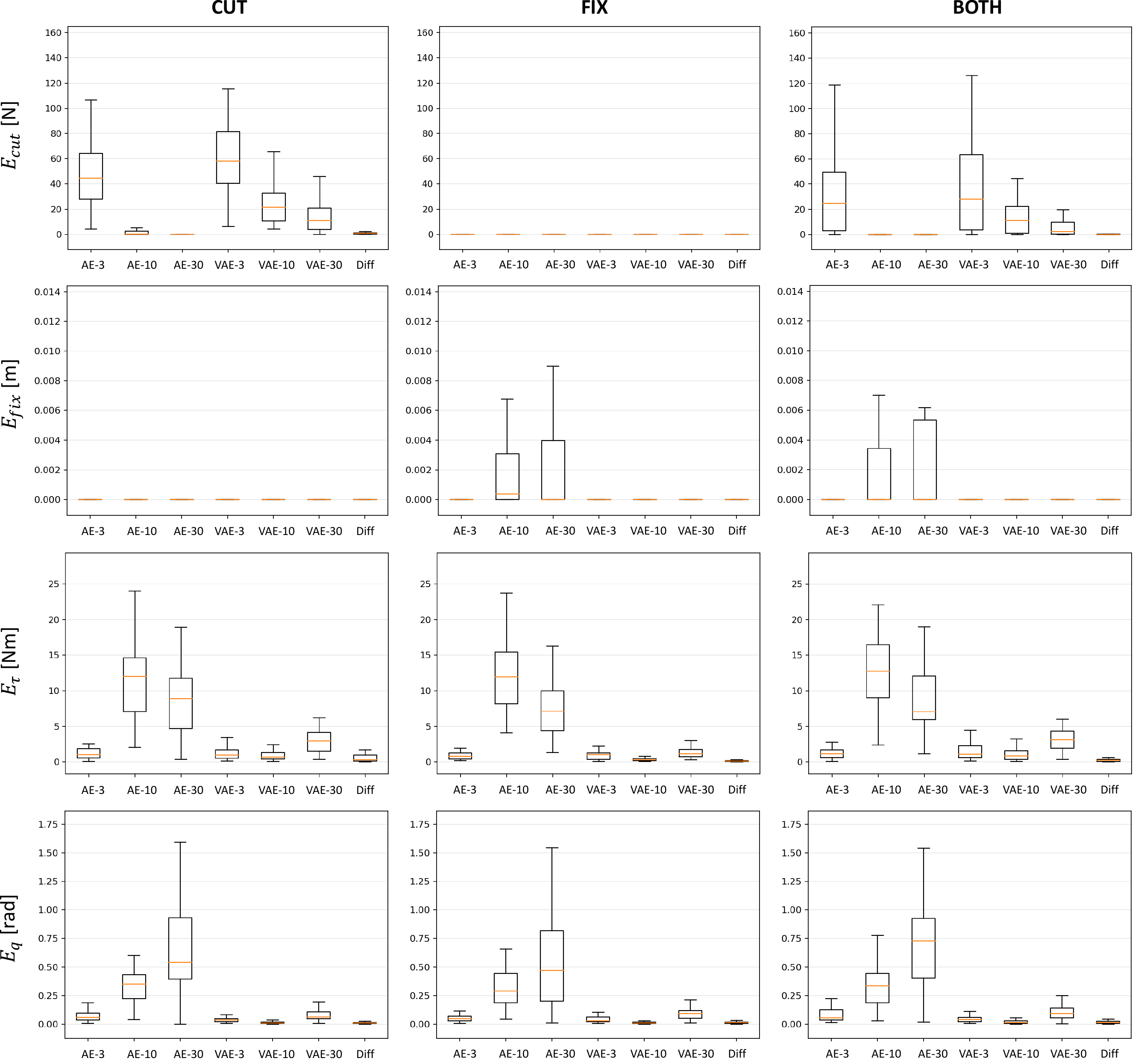}
  \vspace{-1.0ex}
  \caption{Experimental results of adaptation performance under each abnormal condition. The box-and-whisker plots show the distribution of adaptation results over 40 abnormal condition samples for each evaluation metric $E_{cut}$, $E_{fix}$, $E_{\tau}$, and $E_{q}$.}
  \label{figure:exp-results}
  \vspace{-3.0ex}
\end{figure*}

\begin{figure}[t]
  \centering
  \includegraphics[width=0.65\columnwidth]{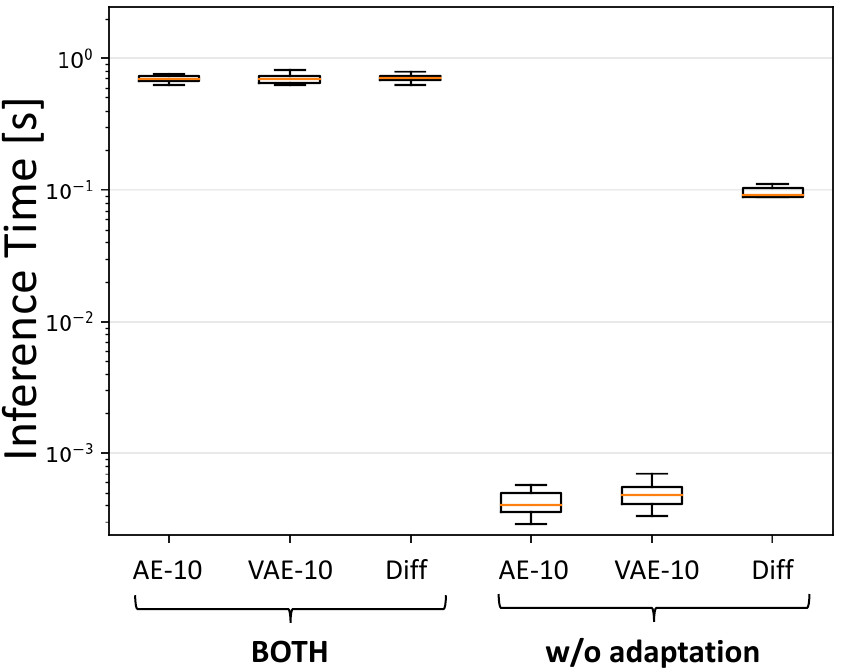}
  \vspace{-1.0ex}
  \caption{Inference time comparison with and without adaptation.}
  \label{figure:exp-time}
  \vspace{-3.0ex}
\end{figure}

\section{Experimental Results and Discussion} \label{sec:experiment}

\subsection{Experimental Results} \label{sec:exp-results}
\switchlanguage%
{%
  We compared the adaptation performance of each model \{autoencoder (AE), variational autoencoder (VAE), diffusion model (Diff)\} under each abnormal condition \{\textbf{CUT}, \textbf{FIX}, \textbf{BOTH}\}.
  The latent space dimensionalities of the AE and VAE are compared for values of 3, 10, and 30, and the results are reported as AE-3, AE-10, AE-30, VAE-3, VAE-10, VAE-30, and Diff, respectively.
  The experimental results are shown in \figref{figure:exp-results}.
  For each evaluation metric $E_{cut}$, $E_{fix}$, $E_{\tau}$, and $E_{q}$, adaptation results over 40 abnormal condition samples are presented using box-and-whisker plots.
  Also, \figref{figure:exp-time} compares the inference times of AE-10, VAE-10, and Diff under the \textbf{BOTH} condition and under a condition without any adaptation.
  When no adaptation is performed, inference in AE and VAE consists only of \equref{eq:ae-vae}, whereas inference in Diff consists solely of the iterative application of \equref{eq:diff-1} and \equref{eq:diff-2}.
  Note that all inference times were measured on an Intel Core i7-10750H CPU and an NVIDIA Quadro T2000 Max-Q GPU.

  First, for the tension error under muscle rupture $E_{cut}$, both under \textbf{CUT} and \textbf{BOTH} conditions, AE-10, AE-30, and Diff exhibit high adaptation performance.
  Through the latent space optimization or the denoising process, the muscle tension of ruptured muscles is successfully driven close to zero.
  In contrast, AE-3 and VAE-based methods in general fail to sufficiently reduce muscle tension, resulting in inferior adaptation performance.

  Next, for the muscle length error under actuator jamming $E_{fix}$, under both \textbf{FIX} and \textbf{BOTH} conditions, all models except AE-10 and AE-30 show high adaptation performance.
  Unlike the case of $E_{cut}$, VAE-based methods exhibit better adaptation performance than AE-based methods.

  Next, for the static equilibrium error $E_{\tau}$, under all abnormal conditions \textbf{CUT}, \textbf{FIX}, and \textbf{BOTH}, AE-3, VAE-3, VAE-10, and Diff demonstrate high adaptation performance.
  In particular, Diff achieves superior adaptation performance.

  Next, for the joint angle error $E_{q}$, under all abnormal conditions \textbf{CUT}, \textbf{FIX}, and \textbf{BOTH}, AE-3, VAE-3, VAE-10, and Diff show high adaptation performance.
  In particular, VAE-10 and Diff achieve the best adaptation performance.

  Finally, regarding inference time, when no adaptation is performed, AE and VAE complete inference within 0.001 s, whereas Diff requires approximately 0.1 s.
  In contrast, when adaptation is enabled, the inference times of AE, VAE, and Diff are all approximately 0.7 s.
}%
{%
  各異常状態\{\textbf{CUT}, \textbf{FIX}, \textbf{BOTH}\}に対して, 各モデル\{AutoEncoder (AE), Variational AutoEncoder (VAE), Diffusion Model (Diff)\}の適応性能を比較した.
  AEとVAEの潜在空間の次元数は3, 10, 30の場合で比較を行っており, それぞれAE-3, AE-10, AE-30, VAE-3, VAE-10, VAE-30, Diffとして結果を示している.
  その実験結果を\figref{figure:exp-results}に示す.
  各評価指標$E_{cut}$, $E_{fix}$, $E_{\tau}$, $E_{q}$に対する40の異常状態サンプルでの適応結果を箱ひげ図で示している.
  また, AE-10, VAE-10, Diffについて, \textbf{BOTH}の条件と一切の適応を行わない条件に対する推論時間の比較を\figref{figure:exp-time}に示す.
  なお, 適応を行わない場合, AEやVAEでは\equref{eq:ae-vae}のみ, Diffでは\equref{eq:diff-1}と\equref{eq:diff-2}の繰り返しのみで推論を行う.

  まず, 筋切断の張力誤差$E_{cut}$に関しては, \textbf{CUT}/\textbf{BOTH}ともに, AE-10, AE-30, Diffが良好な適応性能を示している.
  潜在空間, またはノイズ除去過程により, 切れた筋の張力を0に近づけることができている.
  これに対して, AE-3やVAE全般は, 十分に筋張力を下げることができておらず, 適応性能が劣っている.

  次に, 筋固着の筋長誤差$E_{fix}$に関しては, \textbf{FIX}/\textbf{BOTH}ともに, AE-10, AE-30以外が良好な適応性能を示している.
  $E_{cut}$と異なり, AEに比べてVAEの方が良好な適応性能を示している.

  次に, 静力学バランス誤差$E_{\tau}$に関しては, \textbf{CUT}/\textbf{FIX}/\textbf{BOTH}いずれの異常状態においても, AE-3, VAE-3, VAE-10, Diffが良好な適応性能を示している.
  特にDiffの適応性能が優れていることがわかる.

  次に, 関節角度誤差$E_{q}$に関しては, \textbf{CUT}/\textbf{FIX}/\textbf{BOTH}いずれの異常状態においても, AE-3, VAE-3, VAE-10, Diffが良好な適応性能を示している.
  特に, VAE-10とDiffの適応性能が優れていることがわかる.

  最後に, 推論時間については, 適応を行わない場合, AEやVAEでは0.001 s以内, Diffでは0.1 s程度で推論が完了する.
  これに対して, 適応を行う場合, AE, VAE, Diffのいずれも0.7 s程度の推論時間が必要である.
}%

\subsection{Discussion} \label{sec:subdiscussion}
\switchlanguage%
{%
  Based on the results presented above, we discuss the adaptation performance of each model.
  For the tension error under muscle rupture $E_{cut}$, AEs with larger latent spaces and the diffusion model exhibited high adaptation performance.
  For the muscle length error under actuator jamming $E_{fix}$, VAE-based methods and the diffusion model showed high adaptation performance.
  For the static equilibrium error $E_{\tau}$ and the joint angle error $E_{q}$, AEs and VAEs with smaller latent spaces, as well as the diffusion model, demonstrated high adaptation performance.
  From these results, it can be observed that the diffusion model consistently achieves stable and strong adaptation performance across all evaluation metrics.
  It successfully satisfies the constraints imposed by muscle rupture and actuator jamming, static equilibrium conditions, and the target joint angles.

  In contrast, AE-based methods fail to sufficiently satisfy either the muscle rupture or actuator jamming constraints, and their adaptation performance in terms of static equilibrium and joint angle accuracy is also limited.
  VAE-based methods show high adaptation performance for actuator jamming, but tend to perform poorly under muscle rupture conditions.
  For static equilibrium and joint angle errors, VAEs with a latent space dimensionality of 10 exhibit better adaptation performance than those with latent dimensions of 3 or 30.
  This suggests that selecting an appropriate latent space dimensionality is critical for achieving high adaptation performance with VAEs.

  These results can be explained by several factors.
  To satisfy the constraints imposed by muscle rupture and actuator jamming, the body schema must be capable of representing these constraints.
  Since the latent space of AE-based methods is learned without explicit regularization, optimization can drive the latent variables to values that deviate significantly from the training distribution.
  This often leads to a dilemma in which satisfying the muscle rupture constraint prevents satisfying the actuator jamming constraint, or vice versa.
  In contrast, VAEs impose regularization on the latent space, and increasing the latent dimensionality improves representational capacity, making it easier to satisfy actuator jamming constraints.
  However, when the latent space becomes excessively large, adaptation performance for static equilibrium and joint angle accuracy deteriorates, indicating that careful selection of the latent dimensionality is required.
  Aggregating information into a Gaussian-distributed low-dimensional latent space is therefore likely to hinder stable adaptation to diverse abnormal conditions.

  By contrast, diffusion models can represent diverse data distributions through the denoising process without relying on the size of a latent space.
  As a result, diffusion models can satisfy muscle rupture and actuator jamming constraints while simultaneously achieving high adaptation performance in static equilibrium and joint angle accuracy.
  Rather than compressing information into a latent space, directly utilizing gradient information during the denoising process contributes to stable adaptation under special and abnormal conditions.
  These results indicate that directly operating information in the high-dimensional space, without collapsing or distorting the original data, is effective for robots with redundant sensors and actuators.

  Finally, we discuss the limitations and future challenges.
  This study is limited to simulation experiments.
  The primary reason is that reproducing a wide variety of abnormal conditions on real hardware is difficult, making quantitative evaluation challenging.
  Note that, in musculoskeletal systems, joints and muscles are typically grouped as in \cite{kawaharazuka2021grouping}, with each group usually consisting of no more than approximately five joints and ten muscles.
  Since this is not substantially larger than the two-joint, six-muscle system considered in this study, the network complexity is not expected to differ significantly between simulation and real-robot applications.
  In addition, while this study focuses on muscle rupture and actuator jamming, real robotic systems may experience circuit failures, sensor noise, and various other malfunctions.
  In future work, it will be necessary to investigate which special conditions are likely to occur in practice and to demonstrate the proposed approach in a more practical setting.
  Furthermore, the proposed implementation requires approximately 0.7 s for adaptation, which is acceptable for static motions but insufficient for dynamic behaviors.
  Reducing the inference time will therefore be necessary, and recent advances in diffusion model acceleration may help address this limitation.
  Although this study focuses on musculoskeletal structures, the proposed methodology can be generally applied to robots with redundant sensors and actuators.
  Robots with redundant joints, redundant muscles, or redundant sensors represent a broad range of potential applications.
  We expect that the proposed method will further unlock the adaptive capabilities enabled by redundancy.
}%
{%
  これまでの結果を踏まえ, 各モデルの適応性能について議論する.
  筋切断の張力誤差$E_{cut}$に関しては潜在空間の大きなAEまたはDiffが良好な適応性能を示した.
  また, 筋固着の筋長誤差$E_{fix}$に関してはVAEまたはDiffが良好な適応性能を示した.
  また, 静力学バランス誤差$E_{\tau}$および関節角度誤差$E_{q}$に関しては, 潜在空間の小さなAEまたはVAE, そしてDiffが良好な適応性能を示した.
  これらの結果から, 全ての評価指標において拡散モデルが安定して良好な適応性能を示していることがわかる.
  筋破断や筋固着の制約, 静力学的な制約を満たしつつ, 目標関節角度を実現することに成功している.
  その一方で, AEは筋破断・筋固着のいずれかに対してその制約が全く満たせておらず, 静力学バランスや関節角度の適応性能も高くない.
  VAEは筋固着に対しては良好な適応性能を示すものの, 筋破断に対しては適応性能が低い傾向が見られた.
  静力学バランスや関節角度に関しては, 潜在空間次元が3, 30の場合に比べて, 10の場合に良好な適応性能を示す傾向が見られた.
  つまり, 潜在空間の次元数が適切であることが, VAEの適応性能に重要であることが示唆された.

  これらの結果はいくつかの要因によって説明できると考えられる.
  まず, 筋破断や筋固着の制約を満たすためには, 身体図式がそれらの制約を表現できる必要がある.
  AEの潜在空間は特に制約なく学習されるため, 最適化の結果, 学習時とは大きく異なる値に収束し, 筋破断制約を満たすと筋固着制約が満たせない, あるいはその逆のようなジレンマに陥りやすいと考えられる.
  一方で, VAEは潜在空間に正則化がかかるため, 潜在空間を大きくすることで制約を満たす表現力が向上し, 筋固着制約を満たしやすくなると考えられる.
  しかし, 潜在空間を大きくしすぎると, 静力学バランスや関節角度の適応性能が下がり, 適切な次元数の選択が重要になると考えられる.
  ガウス分布に従う低次元な潜在空間に情報を集約することが, 多様な異常状態に対する安定した適応を妨げていると考えられる.
  これらに対して, 拡散モデルはノイズ除去過程により, 潜在空間の大きさに依存せずに多様なデータ分布を表現できるため, 筋破断・筋固着の制約を満たしつつ, 静力学バランスや関節角度も良好に適応できると考えられる.
  潜在空間に情報を集約せずに, ノイズ除去過程において勾配の情報を用いることが, 特殊な条件下での安定した適応に寄与していると考えられる.
  元々のデータを潰したり歪めたりせずに, 高次元空間で直接扱うことが, 冗長なセンサ・アクチュエータを持つロボットにとって有効であることを示している.

  最後に, 今後の課題について述べる.
  本研究はシミュレーション実験にとどまっている.
  これは, 実機において多様な異常状態を再現することが難しく, 定量的な評価が出来ないことが主な理由である.
  なお, 筋骨格系では大抵関節と筋肉を\cite{kawaharazuka2021grouping}のようにgroupingするため, 最大でも5自由度10筋くらいのことが多く, 今回の2自由度6筋のシミュレーションから大きく外れるものではないため, 実機とシミュレーションでネットワークの複雑度が大きく異なることはないと考えている.
  また, 今回は筋破断とアクチュエータ固着を扱ったが, 実際には回路の不調やセンサのノイズ, その他様々な不具合が考えられる.
  今後, どのような特殊状況が実際にあり得るのかを検証し, より実用的な形で示していく必要があると考えている.
  また, 推論時間について, 提案した実装では適応に0.7 s程度かかっており, 静的な動作としては問題ないものの, 動的な動作に対しては十分な速度とは言えない.
  動的な動作に対しても適応できるようにするためには, 推論時間の短縮が必要であり, 現在盛んに開発の進む拡散モデルの高速化が役に立つ可能性があると考えている.
  最後に, 本研究は筋骨格構造に着目したが, この方法論は冗長なセンサ・アクチュエータを持つロボットに一般的に活用できるものである.
  冗長な関節を持つロボット, 冗長な筋肉を持つロボット, 冗長なセンサを持つロボットなど, その応用の幅は広い.
  冗長性を活用した適応能力が, 本手法によりさらに引き出されることを期待する.
}%

\section{CONCLUSION} \label{sec:conclusion}
\switchlanguage%
{%
  In this study, we proposed methods for body schema learning and adaptation to abnormal states in musculoskeletal robots.
  To address the high dimensionality of the sensor-actuator space in musculoskeletal robots, we investigated body schema learning approaches based on autoencoders, variational autoencoders, and diffusion models.
  Furthermore, for abnormal conditions such as muscle rupture and actuator jamming, we proposed body schema adaptation methods based on latent-space optimization and gradient-guided denoising processes.
  These methods were applied to a simulated musculoskeletal robot model, revealing that autoencoders exhibit reduced adaptation capability as the latent space dimensionality increases, that variational autoencoders achieve good adaptation performance when an appropriate latent dimensionality is chosen, and that diffusion models consistently provide robust adaptation performance across all conditions.
  These results indicate that, for robots with redundant sensors and actuators, directly handling their relationships as high-dimensional information, rather than projecting them into a low dimensional latent space, enables gradient-based adaptation to a wide range of situations.
  In future work, we will continue to evaluate the proposed approach on a broader range of robot morphologies and abnormal conditions.
}%
{%
  本研究では, 筋骨格ロボットの身体図式学習と異常状態適応手法を提案した.
  筋骨格ロボットのセンサ・アクチュエータ空間の高次元性に対処するため, オートエンコーダ, 変分オートエンコーダ, 拡散モデルに基づく身体図式学習手法を提案した.
  また, 筋破断とアクチュエータ固着の異常状態に対して, 潜在空間最適化と勾配ガイダンス・ノイズ除去過程に基づく身体図式適応手法を提案した.
  これらをシミュレーション上の筋骨格モデルに適用し, オートエンコーダは潜在空間の次元を大きくすることで適応能力が下がること, 変分オートエンコーダは適切な潜在空間の次元であれば良好な適応能力が得られること, 拡散モデルはどのような状況下でも良好な適応能力が得られることが分かった.
  これは, 冗長なセンサ・アクチュエータを持つロボットにとって, それらの関係性を低次元な潜在空間に落とし込むのではなく, 高次元情報として直接扱ったほうが, 勾配情報から多様な状況に対応できることを示している.
  今後, より多様なロボット形態とそれらの異常状態への適応能力について評価を継続していく.
}%
{
  \bibliographystyle{IEEEtran}
  \bibliography{main}
}

\end{document}